\documentclass[letterpaper, 10 pt, conference]{ieeeconf}  

\IEEEoverridecommandlockouts                              

\usepackage{graphics} 
\usepackage{epsfig} 
\usepackage{mathptmx} 
\usepackage{times} 
\usepackage{amsmath} 
\usepackage{amssymb}  
\usepackage{booktabs} 

\usepackage{multirow} 
\usepackage{algorithm}   
\usepackage{algorithmic} 

\title{\LARGE \bf
Breaking the Performance-Resource Dilemma: A Lightweight Adaptive Video Inference Enhancement
}

\author{
	Wei Ma, \IEEEmembership{Student Member, IEEE},
	Shaowu Chen,
	Junjie Ye, \IEEEmembership{Student Member, IEEE}, Wei Zhang, \\Tianyao Long,
	 Peichang Zhang$^{*}$, \IEEEmembership{Member, IEEE}, and 
	 Lei Huang$^{*}$, \IEEEmembership{Senior Member, IEEE}
	\thanks{*Corresponding authors}
	\thanks{Wei Ma, Junjie Ye, Peichang Zhang, Lei Huang are with the State Key Laboratory of Radio Frequency Heterogeneous Integration (Shenzhen University), Shenzhen University, Shenzhen 518060, china. (e-mails: 2453044001@mails.szu.edu.cn, 2152432003@email.szu.edu.cn, pzhang@szu.edu.cn, lhuang@szu.edu.cn)}
	\thanks{Shaowu Chen is with the Institute of Applied Artificial Intelligence of the Guangdong–HongKong–Macao Greater Bay, Shenzhen Polytechnic University	Shenzhen, China (email: shaowu-chen@foxmail.com)}
	\thanks{Wei Zhang is with Henan Academy of Science Applied Physics Institute Co.,Ltd. Zhenzhou, China (email: shaowu-chen@foxmail.com)}
}

\begin{document}

\maketitle
\thispagestyle{empty}
\pagestyle{empty}

\begin{abstract}
Video inference (VI) plays a critical role in applications such as autonomous driving and intelligent transportation systems. Existing VI enhancement methods predominantly focus on scaling up model sizes and adopting increasingly sophisticated network architectures. While these approaches have achieved state-of-the-art (SOTA) performance, they often overlook the important  trade-off between inference performance and resource efficiency. This oversight leads to excessive computational demands and restricted deployment flexibility across diverse hardware platforms.To address these limitations, we draw inspiration from the Mixture-of-Experts (MoE) model and propose a fuzzy controller (FC-r) based on key system parameters and inference-related metrics. Guided by the FC-r, we introduce a general and lightweight VI enhancement framework that exploits the spatiotemporal correlations of targets across adjacent video frames. Under real-time resource constraints of the target device, the framework intelligently switches between models of varying scales during the inference process.Experimental results demonstrate that the proposed method effectively achieves a desirable balance between resource utilization and inference performance on different platforms, thereby offering improved generality and practicality for real-world deployment.
\end{abstract}


\section{INTRODUCTION}
\label{sec:intro}

With the rapid integration of artificial intelligence into daily life, video inference (VI) has been widely applied in various domains such as autonomous driving \cite{AutoDrive,IV}, video surveillance \cite{Video_survi}, and traffic flow monitoring \cite{traffic}. Numerous VI methods have been proposed to address various challenges in the VI process, achieving promising results.  For instance, \cite{VCOS}  attained  state-of-the-art (SOTA) performance in zero-shot video segmentation by incorporating optical flow fusion and an open-vocabulary detection branch, while another study \cite{HMA} integrated multiple attention modules and a regional masking mechanism for enhanced video detection performance. In these studies, the inference performance has been primarily boosted by scaling up model sizes \cite{Lenna,SMC1}. Although these methods are effective in improving VI, the resulting large-scale models entail increased parameters and complexity, rendering deployment on resource-constrained platforms particularly challenging \cite{switch}.

To alleviate the inherent trade-off between resource constraints and VI performance improvement, model compression \cite{compression,POCKET,WHC} and hardware-oriented acceleration techniques \cite{FPGA,TRT} have been proposed. However, the former often sacrifices inference performance leading to notable accuracy degradation, while the latter is typically constrained by specific hardware acceleration frameworks, thereby limiting generalizability across platforms.

Recently, adaptive inference mechanisms have increasingly received considerable attention. For example, Kulkarni et al. introduced the concept of a machine learning model balancer and constructed an adaptive machine learning-enabled system, thereby effectively improving the overall quality of service compared with single-model approaches \cite{QoS}. Furthermore, Matathammal et al. adopted an epsilon-greedy strategy to implement CPU-based adaptive model switching on smartphone platforms, thereby achieving a favorable balance between computational load and inference efficiency \cite{EdgeMLBalancer}. In addition, methods such as modular software design \cite{switch} and adaptive system architectures \cite{self-adapt} have also shown potential in addressing the conflict between resource utilization and inference performance. However, most of these methods are designed for cloud systems or still incur considerable computational overhead during runtime, thereby deviating to a certain extent from the original intention of improving resource efficiency.
Mixture-of-Experts (MoE) models have recently gained significant attention due to their considerable application potential in large language models \cite{LLMMoE}, vision models \cite{VideoMoE}, multimodal models \cite{MoE}, and related domains. By dynamically routing inputs to a sparse subset of specialized expert sub-networks instead of activating all parameters, MoE models deliver superior computational efficiency .

\begin{figure*}[!h]
	\centering
	\includegraphics[width=1\textwidth]{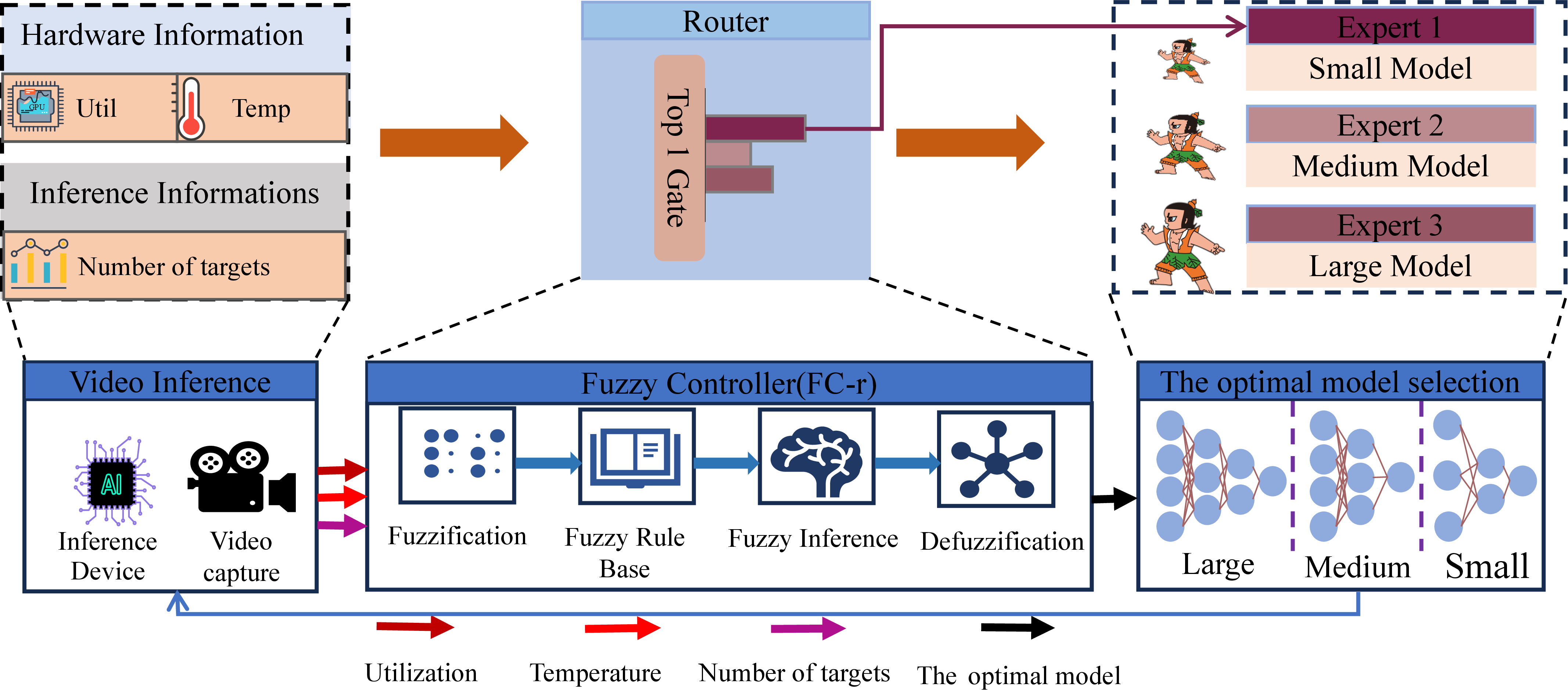}
	\caption{The overall architecture of the proposed framework consists of three main components: the data acquisition module responsible for collecting system status and inference information (on the left), the fuzzy control-based expert router (in the middle), and expert models of varying scales (on the right).}
	\label{Strcture}
\end{figure*}

Inspired by the MoE mechanism, this paper proposes a lightweight VI enhancement framework based on fuzzy controller (FC-r). Notably, whereas  conventional MoE architectures activate sparse sub-networks within a single large model, the proposed approach performs dynamic routing selection among multiple pre-trained, independent models of varying scales. This design choice reduces implementation complexity and makes the method more suitable for heterogeneous edge computing platforms. The overall architecture is illustrated in Fig.~\ref{Strcture}.

The main contributions of this work are summarized as follows:
\begin{itemize}
	\item A FC-r is developed to provide accurate assessments of the inference resource status and to support route selection, which is based on key VI parameters-namely device utilization, device temperature, and the number of targets in the current frame.
	\item Leveraging the spatiotemporal correlation of targets across consecutive video frames, we introduce a dynamic fuzzy switching strategy that enables real-time selection of the optimal inference model based on both the current video context and resource conditions.
	\item Experimental results demonstrate that the proposed method achieves a desirable balance between resource utilization and inference performance on public datasets. 
	\vspace{-0.5em}
\end{itemize}

\section{METHODOLOGY}
\label{sec:method}
Fuzzy control (FC) is an intelligent control paradigm that emulates human-like reasoning and decision-making using fuzzy logic \cite{FC}. To achieve self-adaptive enhanced VI, we design a FC-r-based VI enhancement framework, which is capable of adapting to both core device and inference states. The main components of the proposed controller are as follows.

\begin{figure}[h]
	\centering
	\includegraphics[width=0.8\linewidth]{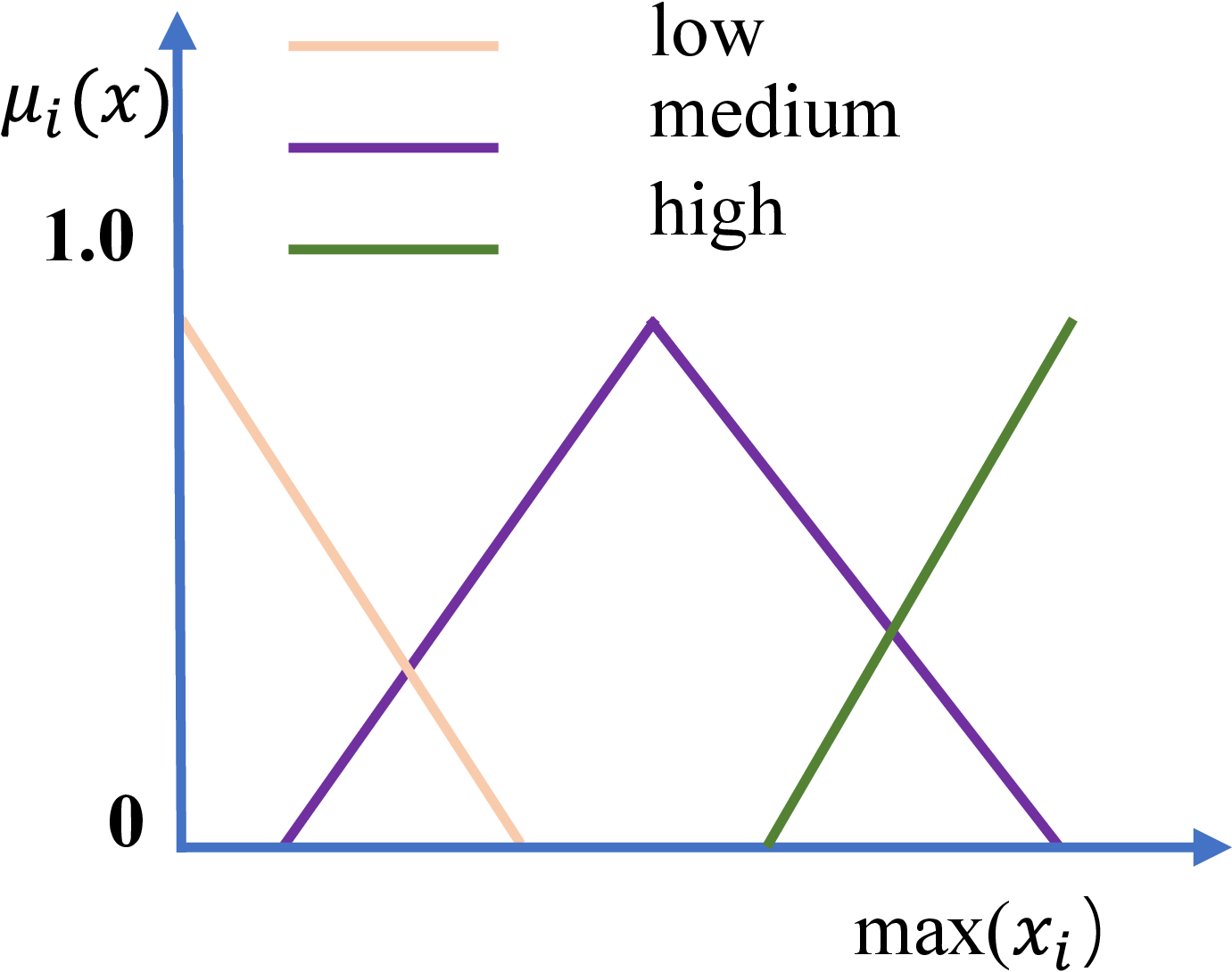}
	\caption{The membership function }
	\label{sec:member}
\end{figure}

\subsection{Fuzzy Controller}
(1) \textbf{Fuzzification: }This module transforms crisp input variables into fuzzy sets using membership functions, as given by
\begin{equation}\label{equa_1}
	\tilde{A} = \left\{ (x, \mu_{\tilde{A}}(x)) \mid x \in U, \mu_{\tilde{A}}(x) \in [0,1] \right\}
\end{equation}
where $U$ denotes the universe of discourse , $\mu_{\tilde{A}}(x)$ represents the membership function, and $\tilde{A}$ denotes the fuzzy set corresponding to that variable. 

In this paper, the triangular membership function is employed, as illustrated in Fig.~\ref{sec:member}. In this study, GPU utilization (GU), GPU temperature (GT), and the number of targets (NT) in the current frame are defined as input variables, while the overall inference score of the current frame is regarded as the output variable. The rationale for selecting these three quantities as fuzzy inputs was to capture the core system and inference states comprehensively. Specifically, GU directly reflects the sufficiency of computational capacity in the inference core, while GT determines whether the system can operate stably over the long term without experiencing frequency throttling or thermal-induced performance degradation. The NT in the current frame directly indicates scene complexity and the potential need for a larger-scale model.

The membership-grade vector of the fuzzified input variables is formulated as 
\begin{equation}\label{equa_2}
	\!\!\boldsymbol{\mu}_i(x_i) \!\!=\!\! \left[ \mu_{\tilde{A}_{i1}}(x_i), \mu_{\tilde{A}_{i2}}(x_i), \dots, \mu_{\tilde{A}_{i{m_i}}}(x_i) \right]^{\text{T}} \!\!\!\!\in\! [0,1]^{m_i}
\end{equation}
where $\boldsymbol{\mu}_i(x_i)$ denotes the fuzzy-result vector of the $i$-th input variable, and its $j$-th component represents the membership degree of the crisp input $x_i$ to the $j$-th fuzzy subset $\tilde{A}_{ij}$ of that variable. Here, $m_i$ represents the total number of fuzzy subsets defined within the universe of discourse of the $i$-th input variable.

Similarly, the fuzzy membership result vector of the output variable is given by
\begin{equation}\label{equa_3}
	\boldsymbol{\mu}_i(y_i) \!=\! \left[ \mu_{\tilde{B}_{i1}}(y_i), \mu_{\tilde{B}{i2}}(y_i), \dots, \mu_{\tilde{B}_{i{l_i}}}(y_i) \right]^{\text{T}} \!\!\in [0,1]^{l_i}
\end{equation}
where $\boldsymbol{\mu}_i(y_i)$ denotes the fuzzy-result vector of the $i$-th output variable, and $l_i$ denotes the total number of fuzzy subsets defined within the $i$-th output variable.

\begin{table}[b]
	\centering
	\caption{Caption of Table of fuzzy inference rules. For GT, GU, and NT, the symbols L, M, and H denote fuzzy subsets corresponding to low, medium, and high degrees, respectively; for the output score selection, the symbols S, M, and L represent the low-, medium-, and high, respectively.}
	\label{tabel_1}
	\footnotesize
	\setlength{\tabcolsep}{3pt}
	\begin{tabular}{cccc|cccc|cccc}
		\toprule
		\textbf{GU} & \textbf{GT} & \textbf{NT} & \textbf{Score} & 
		\textbf{GU} & \textbf{GT} & \textbf{NT} & \textbf{Score} & 
		\textbf{GU} & \textbf{GT} & \textbf{NT} & \textbf{Score} \\
		\midrule
		L & L & L & M & M & L & L & S & H & L & L & S \\
		L & L & M & M & M & L & M & M & H & L & M & S \\
		L & L & H & L & M & L & H & L & H & L & H & M \\
		L & M & L & S & M & M & L & S & H & M & L & S \\
		L & M & M & M & M & M & M & M & H & M & M & S \\
		L & M & H & M & M & M & H & M & H & M & H & M \\
		L & H & L & S & M & H & L & S & H & H & L & S \\
		L & H & M & S & M & H & M & S & H & H & M & S \\
		L & H & H & M & M & H & H & M & H & H & H & S \\
		\bottomrule
	\end{tabular}
\end{table}

(2) \textbf{Fuzzy Rules and Inference:} Fuzzy rules serve as decision-making heuristics derived from practical experience, while fuzzy inference applies these rules by leveraging expert knowledge or empirical data. Specifically, when GU, GT, and NT are all low, the system should switch to a larger model to better utilize computational resources and enhance inference performance, as low utilization reflects underutilized computational capacity and the low temperature also indicates that there is no immediate risk of  thermal-induced performance degradation or device failure. Conversely, when all three quantities are high, the system should switch to a smaller-scale model to reduce thermal stress and mitigate the risk of hardware degradation or failure thermal throttling. The complete set of fuzzy rules is summarized in Table~\ref{tabel_1}, and the fuzzy rule base  $R$ is expressed as
\begin{equation}\label{equa_rule}
	R^{(r)}:\quad
	\text{IF};
	\bigwedge_{i=1}^{p}(x_{i};\text{is};A_{i}^{q_{ri}});
	\text{THEN};(y;\text{is};B^{c_{r}}),
\end{equation}
where $q_{ri}\in{1,2,\dots ,m_i}$, $c_r\in{1,2,\dots ,l_i}$, and $p$ denotes the number of input variables. For each rule, a T-norm operator is employed to aggregate the antecedent conditions and determine the corresponding firing strength, i.e.,
\begin{equation}\label{equa_5}
	\alpha_r = \top\!\Bigl(\mu_{A_1^{q_{r1}}}(x_1^*),\,
	\mu_{A_2^{q_{r2}}}(x_2^*),\,
	\dots,\,
	\mu_{A_p^{q_{rp}}}(x_p^*)\Bigr).
\end{equation}
The firing strength $\alpha_r$ of rule $r$ modulates its consequent fuzzy set $B^{c_r}$.
By applying an $\min$ operator, the output fuzzy set $B_r^\prime$ induced by this rule is
\begin{equation}\label{equa_6}
	\mu_{B_r^\prime}(y)
	= \min \bigl(\alpha_r,\,\mu_{B^{c_r}}(y)\bigr),
	\quad \forall y\in Y.
\end{equation}

The fuzzy sets inferred by all $R$ rules are aggregated into a single overall output fuzzy set by means of a  t-conorm ($\bot$) operator, which is given by
\begin{equation}\label{equa_7}
	\mu_{B^{\ast}}(y)=\bot\!\bigl(\mu_{B_{1}^{\prime}}(y),\,\mu_{B_{2}^{\prime}}(y),\,\dots,\,\mu_{B_{R}^{\prime}}(y)\bigr),~~\forall y\in Y,
\end{equation}

To ensure the intuitiveness of inference results, computational efficiency, and avoidance of information redundancy, this study selects the $\max$ operation from the family of $\bot$  as the aggregation operator:

\begin{equation}
	\mu_{B^{\ast}}(y)=\max_{r=1,\dots,R}\!\bigl(\mu_{B_{r}^{\prime}}(y)\bigr).
\end{equation}

(3)\textbf{ Defuzzification:} Defuzzification is the process of converting the output fuzzy set of the controller into a crisp value, a process typically implemented via membership functions. In this paper, the centroid method is adopted for defuzzification.

\begin{equation} \label{equa_9}
	z^* = \frac{\int_{Y} \mu(y) \cdot y \, dy}{\int_{Y} \mu(y) \, dy}
\end{equation}
where $z^*$ represents the defuzzified output value; $y$ is the value of the output variable within the universe of discourse $Y$; and $\mu(y)$ is the membership function of the output fuzzy set, which quantifies the degree to which the output belongs to the set.

\begin{algorithm}[h]
	\caption{Adaptive Model Selection for VI}
	\label{alg:video-model-selection}
	\begin{algorithmic}[1]
		\REQUIRE Frame Sequence. $F_1,\dots,F_n$; Expert Models $\{M_1,\dots,M_k\}$; Threshold $K$; Fuzzy rules $R$
		\ENSURE Frame-wise adaptive inference results
		
		\STATE \textbf{Init:} $i=0$; $prev\_M=M_1$; $p=0$ (number of processed frames)
		
		\WHILE{$p < n$}
		\STATE Acquire current frame $F_{p}$, extract  features $(GU,GT,NT)$
		
		\STATE \textbf{Fuzzy Inference:} 
		\STATE \quad Compute fuzzy membership vector via (\ref{equa_2})
		\STATE \quad Derive rule firing strengths via  (\ref{equa_5})
		\STATE \quad Perform inference and aggregation via (\ref{equa_6})-(\ref{equa_7})
		
		\STATE \textbf{Model Selection:} 
		\STATE \quad Defuzzify to obtain $M_{opt}$ via (\ref{equa_9});
		\STATE \quad $i=i+1$ if $M_{opt}\neq prev\_M$ else $i=0$
		
		\IF{$i\geq K$} 
		\STATE $prev\_M=M_{opt}$; $i=0$
		\ENDIF
		\STATE Infer $F_{p+1}$ with $prev\_M$

		\STATE Output inferred results
		\ENDWHILE
	\end{algorithmic}
\end{algorithm}

\subsection{Adaptive Model Switching}
The FC-r is employed to dynamically determine the optimal model selection in real time according to the device's resource state and the current inference state. Generally, NT between adjacent video frames does not change abruptly. Therefore, the optimal model determined from the current frame can be effectively applied to the inference of the next frame. Additionally, a model-switching safety mechanism is incorporated to enhance stability. Model switching is triggered only when the same model recommendation is consistently observed across a number of consecutive frames exceeding a predefined safety threshold $K$. This design effectively prevents unnecessary model switching caused by noise or transient errors. The overall algorithm is summarized in \textbf{Algorithm \ref{alg:video-model-selection}}.

\section{Experiment Results}

\subsection{Experiment Setup}

\begin{figure*}[!t]
	\centering
	\includegraphics[width=\textwidth]{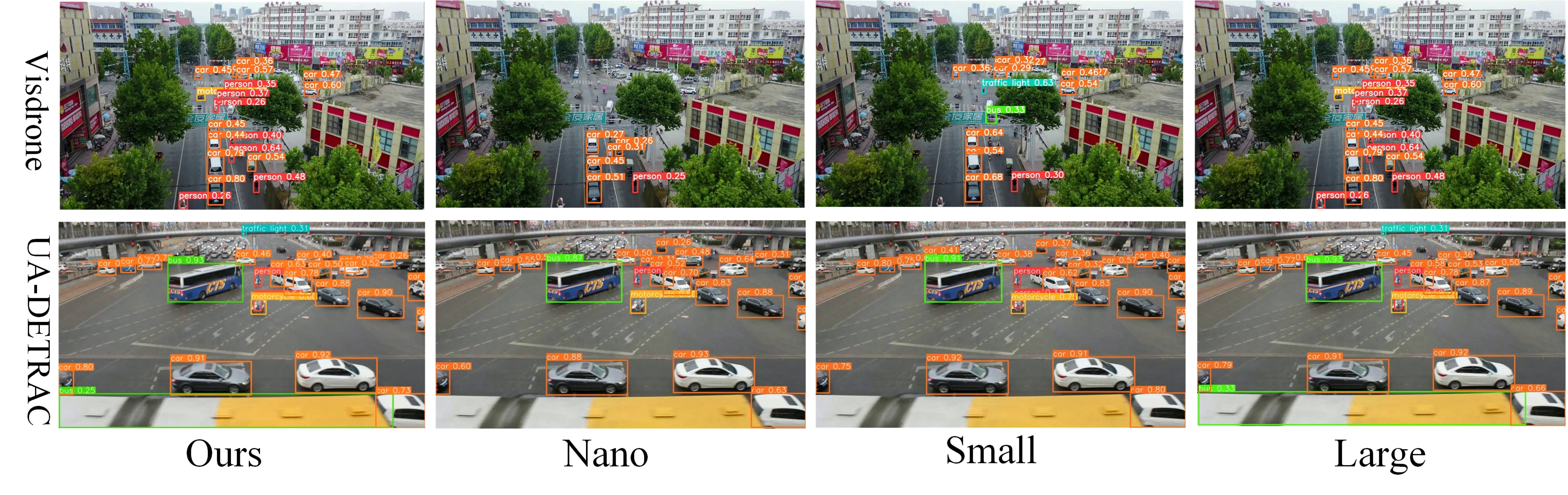}
	\caption{Detection Results on the VisDrone and UA-DETRAC}
	\label{sec:pic1}
\end{figure*}

To evaluate the generalization capability and effectiveness of the proposed algorithm, we conducted comprehensive experiments on both the CNN-based YOLO model family and the Transformer-based DETR model family, while performing a fair comparison with three scale variants (nano, small, and large) of each baseline. It is worth noting that all models across different scales were fine-tuned for 50 epochs on the corresponding datasets to ensure they were well-adapted to the target scenarios represented in the datasets. All evaluation experiments were implemented and validated on two independent hardware platforms, each configured with its own dedicated software environment:
\begin{enumerate}
	\item \textbf{Embedded Platform}: NVIDIA Jetson Orin NX, equipped with a software stack of PyTorch 1.13, CUDA 11.7, and Ubuntu 20.04 LTS operating system.
	\item \textbf{PC Platform}: A PC equipped with an AMD Ryzen 9 9950X CPU, an NVIDIA GeForce RTX 4070 Ti SUPER GPU, 64~GB of RAM, and 2~TB SSD; the software environment is configured with PyTorch 2.0, CUDA 11.8, and Windows 11 operating system.
\end{enumerate}

For evaluation, we adopt selected video sequences from two widely-used public datasets for vision-based detection tasks: the VisDrone\cite{Visdrone} Dataset (for UAV scenarios) , and the UA-DETRAC \cite{WEN2020102907} Dataset (for traffic surveillance). Specifically, we use the VisDrone video sequence (ID: \texttt{uav0000182\_0000\_v}, 363 frames) and two UA-DETRAC video sequences (IDs: \texttt{MVI\_40742}, \texttt{MVI\_40743}, 3284 frames in total).

\subsection{Results}

Table~\ref{P_R} summarizes the precision and recall performance of the proposed method across the VI pipeline. As shown in the table, thanks to its dedicated dynamic adaptive inference mechanism, the proposed method achieves a desirable trade-off between precision and recall while strictly maintaining the upper and lower bounds of detection performance.

\begin{table}[t]
	\centering
	\caption{Results of P, R across Different Platforms and Datasets}
	\label{P_R}
	\begin{tabular}{lccc} 
		\toprule
		\textbf{Platform@dataset} & \textbf{methods} & \textbf{P} & \textbf{R} \\
		\midrule
		& Detr-n & 0.7384 & 0.4406 \\ 
		\multirow{3}{*}{\centering PC@UA-DETRAC} & Detr-s & 0.7892 & 0.5693 \\ 
		& Detr-l & 0.8861 & 0.8207 \\ 
		& Ours & 0.7805 & 0.7581 \\ 
		\midrule 
		\multirow{4}{*}{\centering Jetson@Visdrone} & YOLOv8-n & 0.3641 & 0.3372 \\ 
		& YOLOv8-s & 0.4224 & 0.4188 \\ 
		& YOLOv8-l & 0.499 & 0.4781 \\ 
		& Ours & 0.3952 & 0.3779 \\ 
		\bottomrule
	\end{tabular}
\end{table}

Table~\ref{switch_num} summarizes the variation in the number of model switches across video frames. During frames 501 to 1000, when the traffic signal is red, the number of targets in the video frames remains stable, and the model performs inference smoothly. However, the increase in switching frequency observed between frames 1001 and 1500 is primarily attributed to the traffic light changing from red to green, which triggered vehicle movement at the intersection and caused sustained variations in the number of vehicles. Notably, only 22 switches were recorded throughout the entire 3,284-frames video, which underscores the robustness of the proposed method.

\begin{table}[h]  
	\centering
	\caption{Model Switches in Different Frame Ranges}
	\begin{tabular}{cc}  
		\toprule
		Frames Range & Number of Switches  \\
		\midrule
		0--500        & 3              \\
		501--1000     & 0              \\
		1001--1500    & 10              \\
		1501--3285   & 9             \\
		\bottomrule
	\end{tabular}
	\label{switch_num}
\end{table}

Fig.~\ref{sec:pic1} presents the detection results of the proposed method in comparison with single-model inference on the VisDrone and UA-DETRAC datasets. As shown in the figure, when targets are densely distributed and the device temperature is moderate, the system automatically switches to a larger-scale model for inference. This allows maximizing target detection accuracy and ensuring robust performance in complex scenarios.

To quantify the efficiency of resource utilization, we introduce a resource-performance trade-off metric that comprises the Precision-to-Memory Ratio (PM) and the Recall-to-Memory Ratio (RM). Their definitions are as follows:
\begin{equation}\label{index}
	PM = \frac{\text{Precision}}{\text{Memory Utilization}}.
\end{equation}

\begin{equation}\label{index}
   RM = \frac{\text{Recall}}{\text{Memory Utilization}}.
\end{equation}

\begin{figure}[t]
	\centering
	\includegraphics[width=\linewidth]{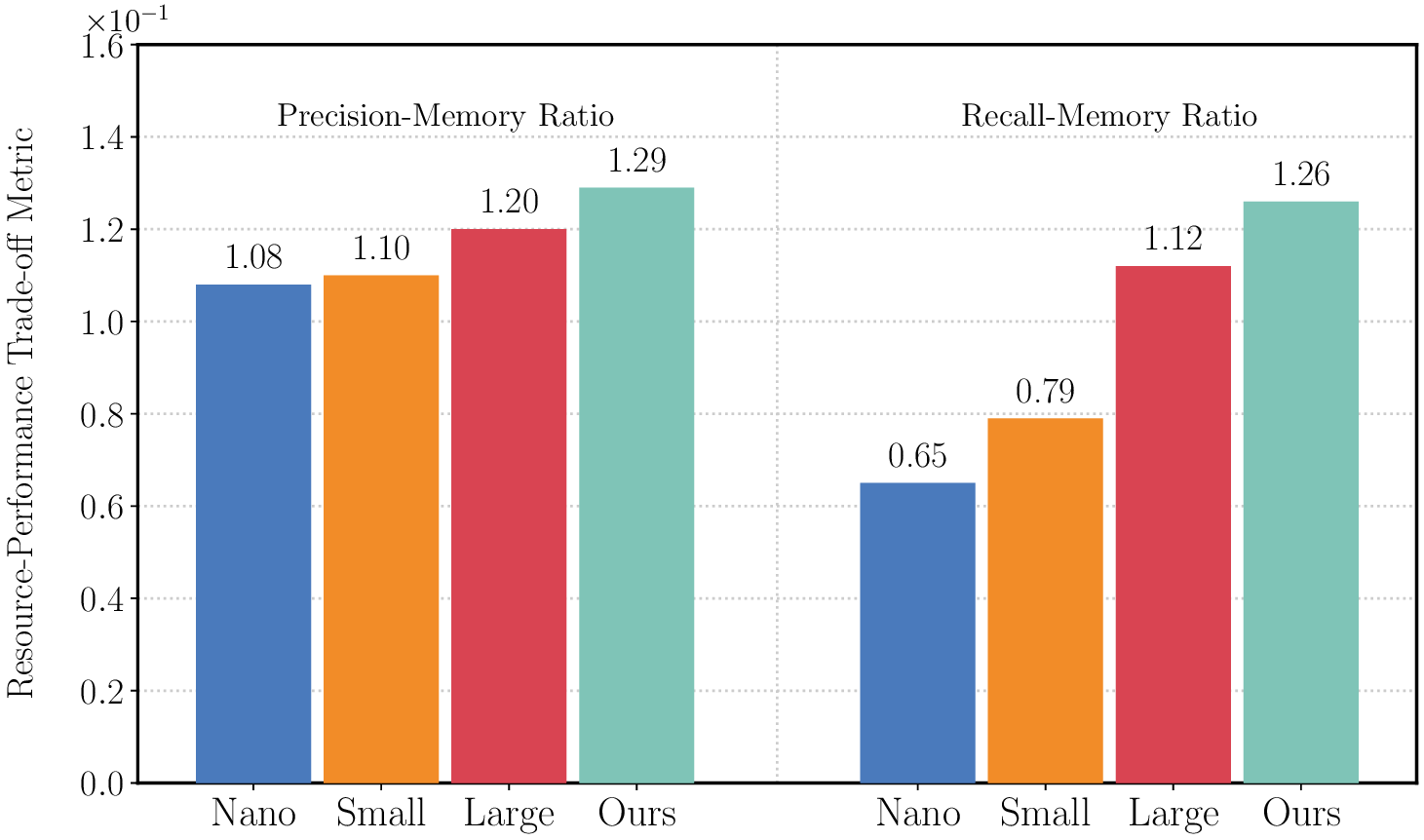}
	\caption{The Value of PM and RM on PC }
	\label{sec:pic2}
\end{figure}

By leveraging dynamic model selection based on per-frame video context and hardware resource conditions, the proposed method achieves a desirable balance between detection performance and resource consumption on both the embedded platform and the PC. Fig. ~\ref{sec:pic2} further illustrates that the proposed algorithm achieves an effective balance between resource utilization and detection performance. Furthermore, the computational complexity of the fuzzy control algorithm is $\mathcal{O}(nm_i+{m_i}^n)$ \cite{algithim_fc}, where $n$ denotes the number of input variables, and $m_i$ represents the number of membership functions defined for the $i$-th variable. Typically, the values of $n$ and $m_i$ are very small; in this study, both are equal to 3, which imposes only a negligible computational burden on the device.

\begin{figure}[tbh]  
	\centering
	\includegraphics[width=1\linewidth]{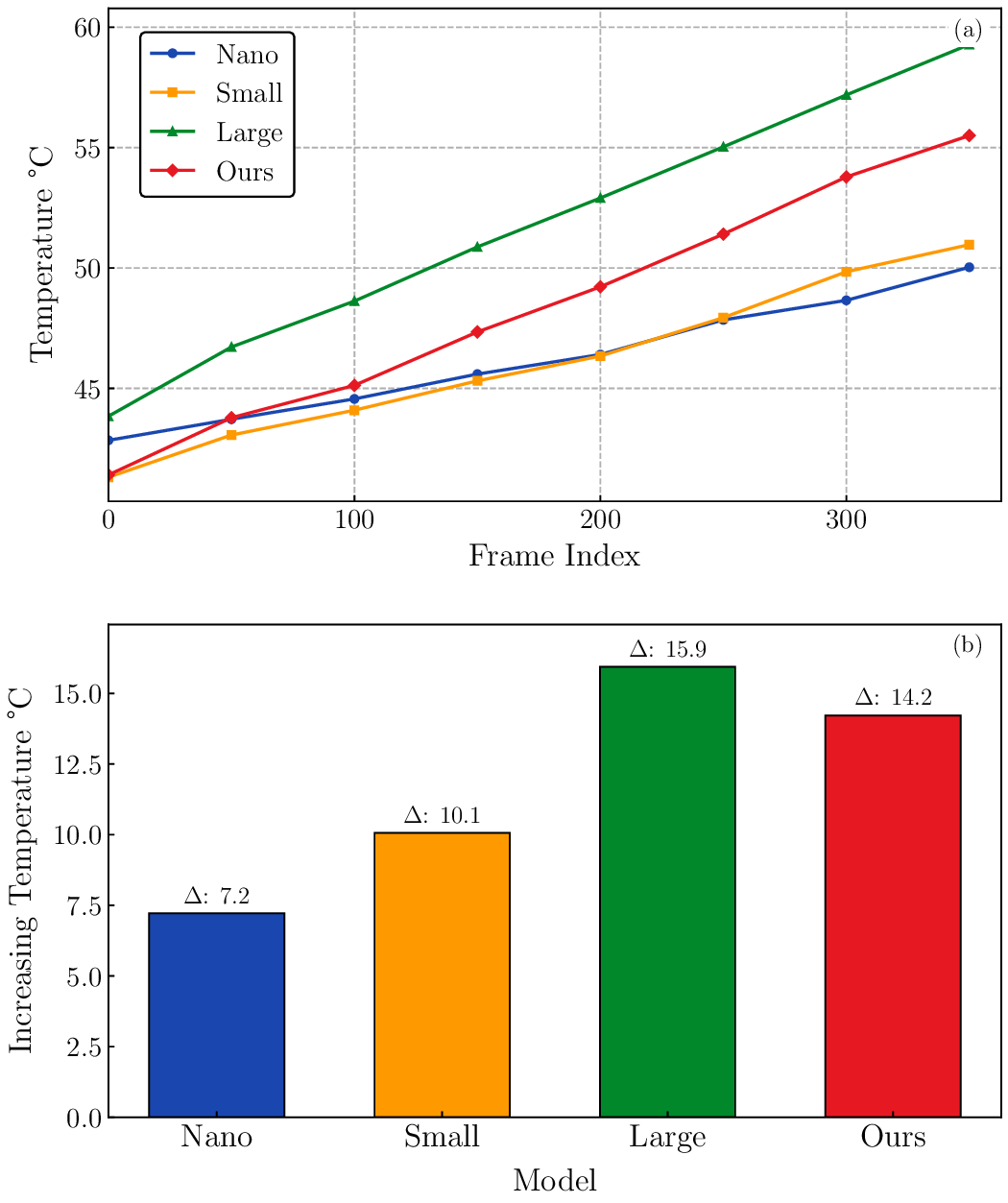}
	\caption{Curves of temperature variation with frame rate for different methods on video stream frames of the UA-DETRAC}
	\label{sec:pic3}
\end{figure}

Fig.~\ref{sec:pic3} presents the core temperature variation (as a proxy for power consumption) of different methods as a function of the number of processed frames in the video stream. In this experiment, inference was performed on a 363-frame video sequence from the UA-DETRAC dataset, during which the core temperature variation of the Jetson Orin NX inference unit was recorded. As shown, the proposed dynamic model selection method effectively suppresses the rapid temperature rise, thereby lowering the risk of device overheating-induced shutdown due to thermal runaway.

\section{CONCLUSION}

This paper proposes a lightweight dynamic VI method based on the developed FC-r, which effectively mitigates the trade-off between resource utilization and inference performance. Experimental results show that its resource utilization efficiency significantly outperforms that of traditional single-model inference approaches. Furthermore, with constant-level complexity, the method imposes a negligible computational burden on resource-constrained devices.

\section*{ACKNOWLEDGMENT}

This project is supported by the Special Fund for the Cultivation of Independent Innovation Achievements of Postgraduate Students at Shenzhen University, Henan Provincial Science and Technology Research Program Joint Fund (No. 235200810049) and the Shenzhen Key Laboratory of Advanced Navigation Technology Fund (No. ZDSYS20150708162521376).

%
%
%
%
%
%
%
%

\bibliographystyle{IEEEtran}
\bibliography{strings,refs}

\end{document}